\pdfoutput=1

\documentclass[11pt]{article}

\usepackage[]{acl}

\usepackage{times}
\usepackage{latexsym}

\usepackage[T1]{fontenc}

\usepackage[utf8]{inputenc}

\usepackage{microtype}
\usepackage{graphicx}
\usepackage{amsmath,amssymb}
\usepackage{subfigure}
\usepackage{multirow}
\usepackage{booktabs}

\title{Confidence Based Bidirectional Global Context Aware Training Framework for Neural Machine Translation}

\author{Chulun Zhou$^{1}$\footnotemark[2], \ Fandong Meng$^{2}$, \ Jie Zhou$^{2}$,\\ \textbf{Min Zhang}$^{3}$\textbf{,} \  \textbf{Hongji Wang}$^{1}$\textbf{,} \ \textbf{Jinsong Su}$^{1,4}$\footnotemark[1]\\ $^{1}$School of Informatics, Xiamen University, Xiamen \\ $^{2}$Pattern Recognition Center, WeChat AI, Tencent Inc, China \\ $^{3}$Harbin Institute of Technology, Shenzhen \ $^{4}$Pengcheng Lab, Shenzhen \\
 {\tt clzhou@stu.xmu.edu.cn \{fandongmeng,withtomzhou\}@tencent.com} \\ {\tt zhangminmt@hotmail.com \{whj,jssu\}@xmu.edu.cn }}

\begin{document}
\maketitle
\renewcommand{\thefootnote}{\fnsymbol{footnote}}
\footnotetext[2]{This work is done when Chulun Zhou was interning at Pattern Recognition Center, WeChat AI, Tencent Inc, China.}
\footnotetext[1]{Corresponding author}
\renewcommand{\thefootnote}{\arabic{footnote}}
\begin{abstract}
Most dominant neural machine translation (NMT) models are restricted to make predictions only according to the local context of preceding words in a left-to-right manner. Although many previous studies try to incorporate global information into NMT models, there still exist limitations on how to effectively exploit bidirectional global context.
In this paper, we propose a \textbf{C}onfidence \textbf{B}ased \textbf{B}idirectional \textbf{G}lobal \textbf{C}ontext \textbf{A}ware (CBBGCA) training framework for NMT, where the NMT model is jointly trained with an auxiliary conditional masked language model (CMLM). The training consists of two stages: (1) multi-task joint training; (2) confidence based knowledge distillation. At the first stage, by sharing encoder parameters, the NMT model is additionally supervised by the signal from the CMLM decoder that contains bidirectional global contexts. Moreover, at the second stage, using the CMLM as teacher, we further pertinently incorporate bidirectional global context to the NMT model on its unconfidently-predicted target words via knowledge distillation.
Experimental results show that our proposed CBBGCA training framework significantly improves the NMT model by +1.02, +1.30 and +0.57 BLEU scores on three large-scale translation datasets, namely WMT'14 English-to-German, WMT'19 Chinese-to-English and WMT'14 English-to-French, respectively.
\end{abstract}

\section{Introduction}
In recent years, Neural Machine Translation (NMT) has achieved great progress and attracted more attention. Most dominant NMT models mainly adopt an encoder-decoder framework \cite{DBLP:conf/nips/SutskeverVL14,DBLP:journals/corr/BahdanauCB14,DBLP:conf/nips/VaswaniSPUJGKP17,meng2019dtmt,DBLP:journals/tacl/SongGZWS19,miao-etal-2021-prevent} with the teacher-forcing strategy \cite{GoodBengCour16} for training. Despite its success, the unidirectional property of teacher-forcing strategy restricts NMT models to only focus on the local context, i.e., the preceding words of the to-be-predicted target word at each decoder step. Apparently, this strategy tends to be limited because word dependencies are always bidirectional involving both preceding and succeeding words on the target side.

To address this issue, many previous researches attempt to exploit global information on the target side \cite{DBLP:conf/naacl/LiuUFS16,DBLP:conf/emnlp/ZhangXSDZ16,DBLP:conf/iclr/SerdyukKSTPB18,DBLP:conf/aaai/ZhangSQLJW18,DBLP:conf/aaai/SuWXLHZ18,DBLP:journals/taslp/ZhangXSL19,DBLP:conf/aaai/ZhangW0L0X19,DBLP:journals/ai/SuZLQYL19,DBLP:journals/tacl/ZhouZZ19,DBLP:journals/ai/ZhangZZZ20}. Typically, they introduce the modelling of target-side global context in the reverse direction by pairing the conventional left-to-right (L2R) NMT model with a right-to-left (R2L) auxiliary model. However, in these methods, the modelling of reverse global context is separate from the local context of preceding words. Thus, they cannot sufficiently encourage the NMT model to exploit bidirectional global context \cite{DBLP:conf/naacl/DevlinCLT19}.
Meanwhile, some of them adopt bidirectional decoding, which often relies on multi-pass decoding or specially customized decoding algorithms \cite{DBLP:conf/naacl/LiuUFS16,DBLP:conf/aaai/ZhangSQLJW18,DBLP:journals/tacl/ZhouZZ19,DBLP:journals/ai/ZhangZZZ20}. 

Another series of studies \cite{DBLP:conf/nips/ConneauL19,DBLP:conf/naacl/EdunovBA19,DBLP:conf/aaai/WengYHCL20,baziotis2020language,DBLP:conf/aaai/YangW0Z00020,DBLP:conf/acl/ChenGCLL20} resort to leveraging target-side bidirectional global context contained in large-scale pre-trained language models (PLM), such as ELMo \cite{DBLP:conf/naacl/PetersNIGCLZ18} and BERT \cite{DBLP:conf/naacl/DevlinCLT19}. These PLMs are normally not bilingual-aware for translation and trained independently of the NMT model. As a special case, \citet{DBLP:conf/acl/ChenGCLL20} design a conditional masked language modelling objective to make BERT aware of source input during the fine-tuning stage. Nevertheless, in these approaches, the pre-trainings of PLMs are independent of NMT models, limiting the potential of model performance.

\begin{figure}[t]
\centering
\scalebox{0.40}[0.40]{\includegraphics{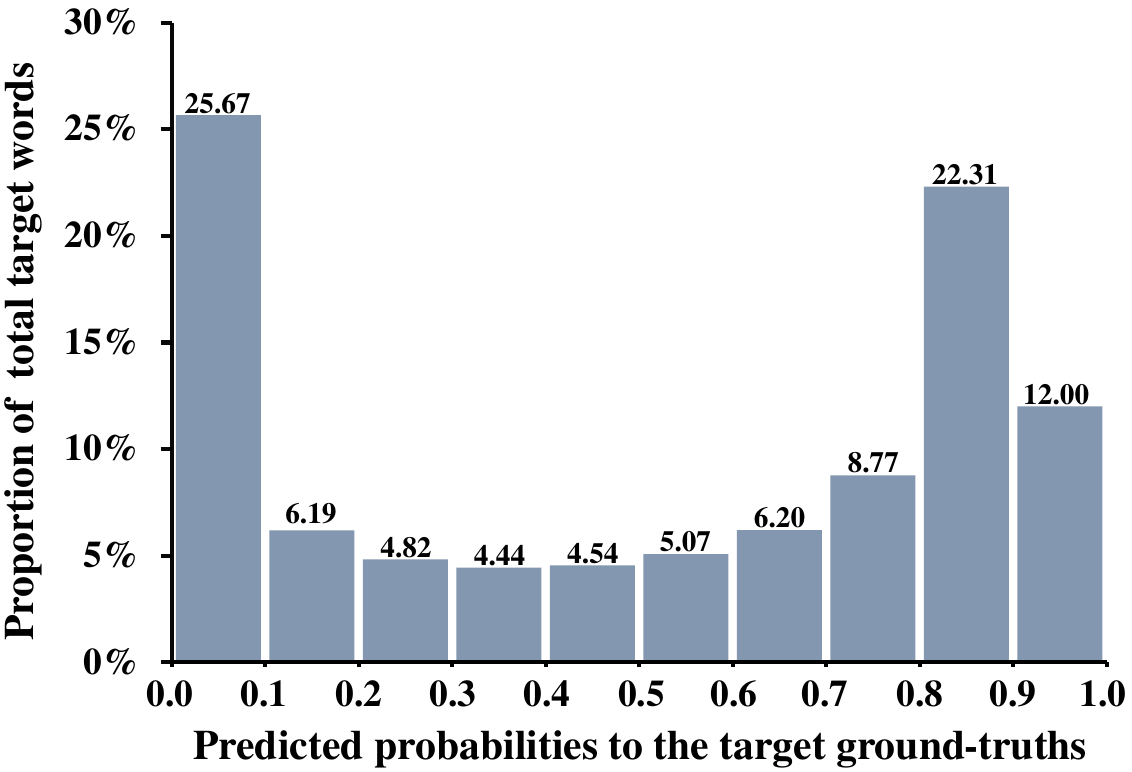}}

\caption{The distribution of the NMT-predicted probabilities to the corresponding ground-truth words on the training set of WMT'14 English-to-German translation task, which is output by a fully-trained Transformer model using teacher-forcing strategy. For instance, the model predicts $25.67\%$ target ground-truth words with probabilities between $0.0$$\sim$$0.1$ given totally correct preceding words at each time step.
}
\label{fig:NMT_probs_on_golden_words}
\vspace{-10pt}
\end{figure}

As for how to effectively incorporate global information into NMT models, another notable deficiency of previous work is that they do not pertinently enhance the NMT model according to its word-level prediction confidence. Ideally, under the teacher-forcing strategy, a well-trained NMT model should assign high probabilities to the target ground-truth words based on correct previous words, which, however, is not the case. Figure~\ref{fig:NMT_probs_on_golden_words} depicts the predicted word-level probabilistic distribution of a fully-trained Transformer model. We find that, even based on totally correct preceding words, there is a considerable portion of target ground-truth words that the model predicts with relatively low probabilities. 
The reasonable cause of this phenomenon is that the NMT model cannot confidently predict these target words according to only the local context of preceding words \cite{DBLP:conf/coling/WatanabeS02,hoangHC17}. Hence, we should especially refine the NMT model on these unconfidently-predicted target words.

In this paper, we propose a \textbf{C}onfidence \textbf{B}ased \textbf{B}idirectional \textbf{G}lobal \textbf{C}ontext \textbf{A}ware (CBBGCA) training framework for NMT. Under our framework, the NMT model is jointly trained with a conditional masked language model (CMLM) which is essentially bilingual-aware and contains bidirectional global context on the target side. Specifically, the CBBGCA training consists of two stages. At the first stage, we jointly train the NMT model and CMLM in a multi-task learning manner by sharing the encoders of the two models. This preliminarily enhances the NMT model because the encoder is additionally supervised by the signal from the CMLM decoder that contains bidirectional global context. At the second stage, we employ the CMLM to pertinently refine the training of the NMT model on those unconfidently-predicted target words via confidence based knowledge distillation. By doing so, our model can be further encouraged to effectively leverage the bilingual-aware bidirectional global context contained in the CMLM.

To sum up, the major contributions of our paper are as follows:
\vspace{-5pt}
\begin{itemize}
\setlength{\itemsep}{5pt}
\setlength{\parsep}{0pt}
\setlength{\parskip}{0pt}
\item
We introduce multi-task learning to benefit the NMT model by sharing its encoder with an auxiliary CMLM, which preliminarily enhances the NMT model to capture bidirectional global context.
\item
We further propose confidence based knowledge distillation using the CMLM as teacher to especially refine the NMT model on unconfidently-predicted target words, more effectively exploiting the bidirectional global contextual information. 
\item 
Extensive experiments on large-scale WMT'14 English-to-German, WMT'19 Chinese-to-English and WMT'14 English-to-French translation tasks show that our CBBGCA training framework respectively improves the state-of-the-art Transformer model by +1.02, +1.30 and +0.57 BLEU points, which demonstrate the effectiveness and generalizability of our approach.
\end{itemize}
\vspace{-10pt}

\section{CBBGCA Training Framework}
\label{section3}
In this section, we will introduce our proposed CBBGCA training framework that employs a CMLM to enhance the NMT model according to its prediction confidence. In the following subsections, we first describe the basic architectures of our NMT model and CMLM. 
Then, we introduce the training procedures of our CBBGCA framework, involving two stages.
\vspace{-5pt}
\subsection{The NMT model and CMLM}
Both the NMT model and CMLM are based on Transformer \cite{DBLP:conf/nips/VaswaniSPUJGKP17}, which is essentially an attentional encoder-decoder framework.\footnote{Please note that our framework can also be adapted to other NMT models.} 

\subsubsection{Encoder}
The encoders of the NMT model and the CMLM are identical, which are mainly used to learn the semantic representations of the source sentence.

Generally, the encoder consists of ${L_{e}}$ identical layers, each of which contains two sub-layers: a self-attention (SelfAtt) sub-layer and a position-wise feed-forward network (FFN) sub-layer. The SelfAtt sub-layer takes the hidden states of the previous layer as inputs and conducts multi-head scaled dot-product attention. Let $\mathbf{h}^{(l)}$ denote the hidden states of the $l$-th encoder layer, the SelfAtt sub-layer can be formulated as
\begin{equation}
    \mathbf{c}^{(l)} = \mathrm{AN}(\mathrm{SelfAtt}(\mathbf{h}^{(l-1)}, \mathbf{h}^{(l-1)}, \mathbf{h}^{(l-1)})),
\end{equation}
where AN($\cdot$) denotes the \textit{AddNorm}, i.e., layer normalization with residual connection. Afterwards, the FFN sub-layer is applied,
\begin{equation}
    \mathbf{h}^{(l)} = \mathrm{AN}(\mathrm{FFN}(\mathbf{c}^{(l)})).
\end{equation}
Note that $\mathbf{h}^{(0)}$ is initialized as the embedding sequence of the source sentence and the hidden states of the $L_{e}$-th layer $\mathbf{h}^{(L_{e})}$ are used as the final word-level representations of the source sentence.

\subsubsection{Decoders}
The decoders of the NMT model and the CMLM are similar except their self-attention mechanisms and prediction manners.
\paragraph{The NMT Decoder.}
It is comprised of $L_{d}$ identical layers with each having three sub-layers: a masked self-attention (MaskSelfAtt) sub-layer, a cross-attention (CrossAtt) sub-layer and an FFN sub-layer. Particularly, to preserve the autoregressive property at each time step, the MaskSelfAtt sub-layer performs self-attention with an attention mask that prevents the decoder from seeing succeeding words. To generate the hidden states $\mathbf{s}^{(l)}$ of the $l$-th decoder layer, the MaskSelfAtt sub-layer can be formulated as 
\begin{equation}
    \mathbf{a}^{(l)} = \mathrm{AN}(\mathrm{MaskSelfAtt}(\mathbf{s}^{(l-1)}, \mathbf{s}^{(l-1)}, \mathbf{s}^{(l-1)})).
\end{equation}
Then, the CrossAtt sub-layer conducts cross-attention using $\mathbf{a}^{(l)}$ and the source representations $\mathbf{h}^{(L_{e})}$,
\begin{equation}
    \mathbf{z}^{(l)} = {\rm{AN}}(\mathrm{CrossAtt}(\mathbf{a}^{(l)}, \mathbf{h}^{(L_{e})}, \mathbf{h}^{(L_{e})})).
\end{equation}
Next, the FFN sub-layer maps $\mathbf{z}^{(l)}$ into $\mathbf{s}^{(l)}$:
\begin{equation}
    \mathbf{s}^{(l)} = \mathrm{AN}(\mathrm{FFN}(\mathbf{z}^{(l)})).
\end{equation}

Finally, with the source sentence $\mathbf{x}$, the target translation $\mathbf{y}_{<t}$ and the learned top-layer hidden states $\mathbf{s}$, the decoder models the probability distribution over the target vocabulary at the $t$-th time step as follows:
\begin{equation}
    p(y_{t}|\mathbf{y}_{<t}, \mathbf{x}) = \mathrm{softmax}(\mathbf{W}\mathbf{s}_{t}),
\label{NMT_prediction}
\end{equation}
where $\mathbf{W}$ represents the learnable parameter matrix for the linear transformation.

\paragraph{The CMLM Decoder.}
Typically, it predicts a set of masked target words $\mathbf{y}_{m}$ given the source sentence $\mathbf{x}$ and the set of observable target words $\mathbf{y}_{o}$. The CMLM decoder also contains $L_{d}$ identical layers, each of which also includes a SelfAtt sub-layer, a CrossAtt sub-layer, and an FFN sublayer. Unlike the MaskSelfAtt sub-layer of the NMT decoder, the attention mask is removed in the SelfAtt sub-layer of the CMLM decoder. 

Finally, with the learned top-layer hidden states $s'$ of the CMLM decoder, the predicted probability distribution for every masked target word $y_{t}\in{\mathbf{y}_{m}}$ can be formalized as
\begin{equation}
    p(y_{t}|\mathbf{y}_{o}, \mathbf{x}) = \mathrm{softmax}(\mathbf{W}'\mathbf{s}'_{t}),
\label{cmlm_prediction}
\end{equation}
where $\mathbf{W}'$ is the learnable parameter matrix of the linear transformation. Note that since the CMLM decoder takes $\mathbf{y}_{o}$ rather than $\mathbf{y}_{<t}$ as input, which includes both preceding and succeeding words with respect to every masked target word, it should contain bidirectional global contextual information.
\subsection{Two-stage Training}
\label{sec:two_stage_training}
The training of CBBGCA framework involves two stages. At the first stage, we jointly train the NMT model and CMLM by multi-task learning. At the second stage, according to the word-level prediction confidence, we employ the CMLM to refine the training of the NMT model through knowledge distillation.
\begin{figure*}[!htb]
\centering
\includegraphics[scale=0.70]{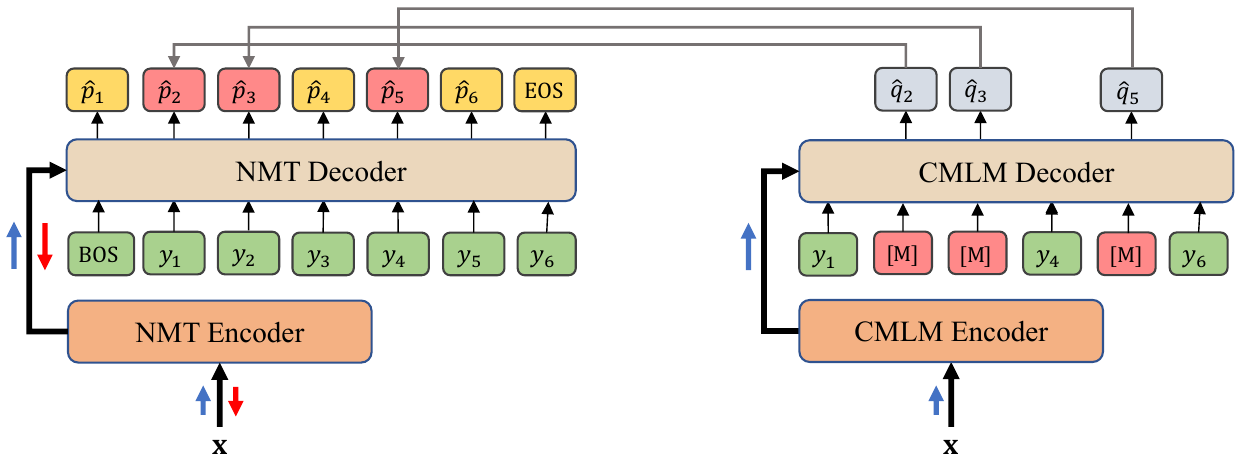}
\caption{
The second training stage. Supposing the NMT-predicted $\hat{p}^{*}_{2}$, $\hat{p}^{*}_{3}$ and $\hat{p}^{*}_{5}$ are lower than $\epsilon$, the set $\mathbf{y}_{m}$ of masked target words becomes $\{y_{2},y_{3},y_{5}\}$. Accordingly, the input $\mathbf{y}_{o}$ to the CMLM decoder is $y_{1},\mathrm{[M],[M]},y_{4},\mathrm{[M]},y_{6}$. The blue and red arrow also represent the forward an backward propagation pass, respectively. Note that we separate the previously-shared encoder and fix the parameters of the CMLM at this stage.
}
\label{fig:stage2}
\vspace{-10pt}
\end{figure*}
\subsubsection{Stage 1: Multi-task Joint Training}
In the first training stage, given a batch of training instances, we jointly train the NMT model and CMLM by simultaneously optimizing their respective objectives:
\begin{equation}
    L_{1}(\theta_{e},\theta_{nd},\theta_{cd}) = {\rm\lambda}L_\mathrm{nmt}\!+\!(1\!-\!\lambda)L_\mathrm{cmlm},
\label{eq:stage1_objective}
\vspace{-5pt}
\end{equation}
where $\lambda$ is a balancing hyper-parameter and $\theta_{e}$, $\theta_{nd}$ and $\theta_{cd}$ denote the parameters of the shared encoder, the NMT decoder and the CMLM decoder, respectively. 
Very importantly, during this procedure of joint training,
we share the same encoder for the NMT model and CMLM. In this manner, the NMT model benefits from the multi-task learning because the encoder is additionally supervised by the signal from the CMLM decoder which contains bidirectional global context. 

Specifically, using the teacher-forcing strategy, the NMT model is optimized through the following objective:
\vspace{-9pt}
\begin{equation}
    L_\mathrm{nmt}(\theta_{e}, \theta_{nd}) = -\sum_{t=1}^{|\mathbf{y}|}{\log p(y_{t}|\mathbf{y}_{<t}, \mathbf{x})}.
\label{eq:nmt_objective}
\vspace{-4pt}
\end{equation}
Besides, we adopt the strategy used in \cite{DBLP:conf/emnlp/GhazvininejadLL19} to optimize the CMLM. 
Concretely, we randomly select $n$ words, where $n$$\sim$$uniform(1,|\mathbf{y}|)$, and replace each word with a special token [M], splitting $\mathbf{y}$ into $\mathbf{y}_{o}$ and $\mathbf{y}_{m}$. Formally, we optimize the CMLM by minimizing the following objective for every word in $\mathbf{y}_{m}$:
\begin{equation}\label{eq:cmlm_objective}
    L_\mathrm{cmlm}(\theta_{e}, \theta_{cd})=-\sum_{y_{t}\in{\mathbf{y}_{m}}}{\log p(y_{t}|\mathbf{y}_{o}, \mathbf{x})}.
\end{equation}

\subsubsection{Stage 2: Confidence Based Knowledge Distillation (CBKD)}
\label{subsubsec:CBKD}
At the second stage, once we obtain the two fully-trained models, we use the CMLM to further refine the training of the NMT model through knowledge distillation (KD). The reason why we introduce such a KD-based model training is that the conventional NMT model predicts a considerable portion of target ground-truth words with relatively low probabilities, as shown in Figure~\ref{fig:NMT_probs_on_golden_words}. This phenomenon indicates that the NMT model cannot confidently predict these target words based on only local context of preceding words. Therefore, we aim to pertinently distill the knowledge of CMLM into the NMT model on these unconfidently-predicted target words because the CMLM contains bilingual-aware bidirectional global context.

Figure~\ref{fig:stage2} depicts the training procedure of this stage with an illustrative example. Given the source sentence $\mathbf{x}$ and the preceding ground-truth words $\mathbf{y}_{<t}$ at each time step $t$, we first let the NMT model make predictions for every target word using Equation~\ref{NMT_prediction}, producing word-level probability distributions $\hat{p}_{1},\hat{p}_{2},...,\hat{p}_{|\mathbf{y}|}$. Then, we determine the word set $\mathbf{y}_{m}$ where the predicted probabilities $\hat{p}^{*}_{t}$ to the corresponding ground-truth words are lower than a threshold value $\epsilon$,
\begin{equation}\label{eq:word_selection}
    \mathbf{y}_{m}=\{y_{t}|\hat{p}^{*}_{t} \le \epsilon,1 \leq t \leq |\mathbf{y}|\}.
    \vspace{-2pt}
\end{equation}
Next, we obtain the set $\mathbf{y}_{o}$ of partially observable target words by replacing those selected ground-truth words with a special token [M]. Subsequently, we feed $\mathbf{y}_{o}$ to the CMLM and obtain its predicted probability distribution $\hat{q}_{t}$ for every word in $\mathbf{y}_{m}$ using Equation~\ref{cmlm_prediction}.

To pertinently refine the NMT model on the set $\mathbf{y}_m$ of its unconfidently-predicted target words, we use the CMLM with fixed parameters as teacher and transfer its knowledge to the NMT model. Along with the supervision from the corresponding ground-truth words, we optimize the NMT model with a balancing factor $\alpha$ through the following objective:
\begin{equation}\label{eq:kd_objective}
    L_{\mathrm{kd}}(\theta_{ne},\!\theta_{nd})\!=\!\sum_{y_{t}\in\mathbf{y}_{m}}{\!\{\!{\rm\alpha{KL}}(\hat{q}_{t}||\hat{p}_{t})}\!-\!{\rm(1\!-\!\alpha)\!}\log\hat{p}^{*}_{t}\!\},
    \vspace{+5pt}
\end{equation}
where $\theta_{ne}$, $\theta_{nd}$ and KL($\cdot$) represent the NMT encoder parameters, the NMT decoder parameters and the Kullback–Leibler divergence \cite{NIPS2015_8d55a249}, respectively. Here, we follow \cite{clark2019bam} to linearly decrease the factor $\alpha$ from 1 to 0 throughout training. This guides the NMT model to absorb more knowledge from the CMLM at the early period of the stage 2 and gradually re-focus on the ground-truth words to learn better. Finally, the total training objective of this stage is as follows:
\begin{equation}
    L_{2}(\theta_{ne},\theta_{nd}) = L_{\mathrm{kd}}(\theta_{ne},\!\theta_{nd})-\!\sum_{y_t\in\mathbf{y}_{o}\backslash\mathrm{[M]}}{\!\log \hat{p}^{*}_{t}},
\label{eq:stage2_objective}
\end{equation}
where $\mathbf{y}_{o}\backslash{\mathrm{[M]}}$ represents $\mathbf{y}_{o}$ excluding all special tokens [M].
By doing so, we can fully strengthen the ability of the NMT model to leverage the bilingual-aware bidirectional global context contained in the CMLM. Note that the CMLM is not involved at the inference time.
\vspace{0pt}
\section{Experiments}
\vspace{0pt}
\subsection{Datasets}
We carry out experiments on three large-scale translation tasks, WMT'14 English-to-German (En${\rightarrow}$De), WMT'19 Chinese-to-English (Zh${\rightarrow}$En) and WMT'14 English-to-French (En${\rightarrow}$Fr). The data are preprocessed using Byte-Pair-Encoding\footnote{https://github.com/rsennrich/subword-nmt} \cite{DBLP:conf/acl/SennrichHB16a} (BPE). More dataset statistics and the detailed preprocessing procedures are described in Appendix~\ref{sec:appendix_A}.

\subsection{Implementation Details}
We follow the settings used in \cite{DBLP:conf/nips/VaswaniSPUJGKP17} to build the NMT model under \textit{Transformer-base} configuration. Concretely, the \textit{Transformer-base} architecture is comprised of 6 encoder and decoder layers, each with 512 as hidden size, the FFN sub-layers of 2,048 dimension and 8 heads in multi-head attentions. For more details about the training and inference, please refer to Appendix~\ref{sec:appendix_B}.

\begin{figure}
    \centering     
    \subfigure[]{
        \label{fig:hyper_lambda}
        \begin{minipage}{.45\columnwidth}
        \includegraphics[width=\columnwidth]{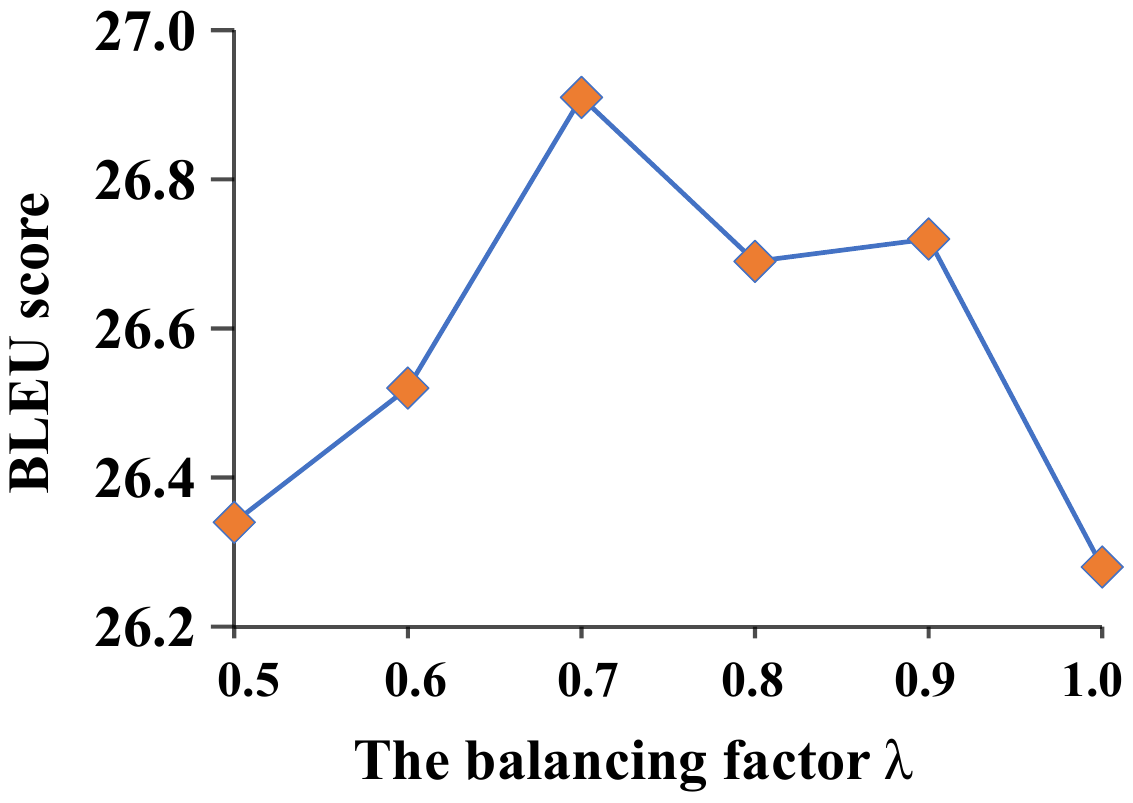}
        \end{minipage}
    }
    \subfigure[]{
        \label{fig:hyper_epsilon}
        \begin{minipage}{.45\columnwidth}
        \includegraphics[width=\columnwidth]{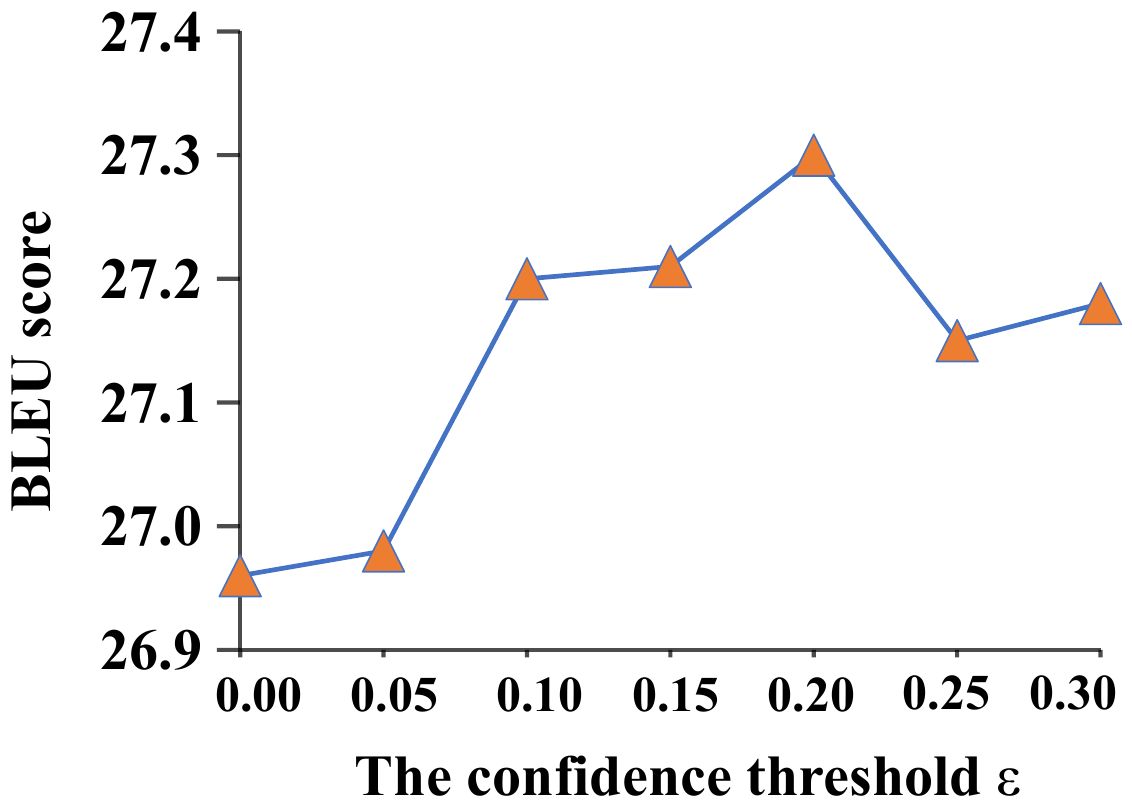}
        \end{minipage}
        }
    \vspace{-5pt}
    \caption{The performance of the NMT model on the WMT'14 $\rm En{\rightarrow}De$ validation set with different hyper-parameters. (a) The BLEU scores after the first stage with different $\lambda$, the balancing factor in Equation~\ref{eq:stage1_objective}.
    (b) The BLEU scores after the second stage with different $\epsilon$, the confidence threshold in Equation~\ref{eq:kd_objective}.}
    \vspace{-10pt}
\end{figure}
\subsection{Hyper-parameters}
Apart from all the hyper-parameters we empirically set based on previous experience, the balancing factor $\lambda$ in Equation~\ref{eq:stage1_objective} and the confidence threshold $\epsilon$ for determining $\mathbf{y}_m$ in Equation~\ref{eq:stage2_objective} are the hyper-parameters we need to manually tune on the validation set.

To balance the training of the NMT model and CMLM, we select the minimum $\lambda$ that can bring steady improvements to the NMT model within 200,000 steps. As shown in Figure~\ref{fig:hyper_lambda}, we gradually vary $\lambda$ from $0.5$ to $1.0$ with an increment of $0.1$ and evaluate the performance on the validation set. We find that the NMT model achieves its peak when $\lambda=0.7$. Hence, $\lambda$ is set to $0.7$ for the joint training of the two models at the first stage.

Given the selected $\lambda$, at the second training stage, we also investigate the impact of $\epsilon$ on the validation set. We adjust its value from $0.0$ to $0.3$ with an interval of $0.05$. As shown in Figure~\ref{fig:hyper_epsilon}, the NMT model performs the best when the $\epsilon$ is $0.2$. Therefore, we set $\epsilon=0.2$ as the confidence threshold for the second training stage.

\subsection{Main Results}
In our experiments, \textbf{CBBGCA} is the system under our proposed training framework, as described in Section~\ref{sec:two_stage_training}. \textbf{Multi-300k} denotes the baseline system that jointly optimizes the NMT model and CMLM by sharing their encoders throughout the whole 300k training steps, which is used to make comparison with conducting CBKD at the second training stage. For evaluation, in addition to the widely used BLEU \cite{DBLP:conf/acl/PapineniRWZ02}, we also adopt the Comet \cite{DBLP:conf/emnlp/ReiSFL20} which is recently a more welcomed metric.

\begin{table}[!tbp]
    \centering
    \renewcommand{\arraystretch}{1.1}
    \setlength{\tabcolsep}{1.5mm} {
    \scalebox{0.86}{
    \begin{tabular}{ l | l | l }
    \toprule
    \hline
    \textbf{System} & \textbf{BLEU} & \textbf{Comet} \\ 
   \hline
    \multicolumn{3}{c}{ Existing Systems} \\
    \hline
     Transformer \cite{DBLP:conf/nips/VaswaniSPUJGKP17}$^{*}$ & 27.30 & - \\
    \hline 
    Rerank-NMT \cite{DBLP:conf/naacl/LiuUFS16}$^{*}$ &  27.81 & - \\
    ABD-NMT \cite{DBLP:conf/aaai/ZhangSQLJW18}$^{*}$ &  28.22 & - \\
    FKD-NMT \cite{DBLP:journals/taslp/ZhangXSL19}$^{**}$ &  27.84 & - \\
    SB-NMT \cite{DBLP:journals/tacl/ZhouZZ19}$^{*}$ &  \textbf{29.21} & - \\
    DBERT-NMT \cite{DBLP:conf/acl/ChenGCLL20}$^{*}$ &  27.53 & - \\
    \hline
    \multicolumn{3}{c}{Our Systems} \\
    \hline
    Transformer &  27.30 & 0.2602 \\
    Multi-300k & 27.88$\dagger$ & 0.2703$\dagger$ \\
    CBBGCA & \textbf{28.32}$\ddagger\diamond$& \textbf{0.2828}$\ddagger\diamond$ \\
    \hline
    \end{tabular} }}
    \caption{BLEU (\%) and Comet scores on WMT'14 En${\rightarrow}$De. `*': results are taken from corresponding papers. `**': results are reproduced by running original code. `$\dagger$' and `$\ddagger$': significantly better than our Transformer with t-test $p$$<$${0.05}$ and $p$$<$${0.01}$, respectively. `$\diamond$': significantly better than FKD-NMT with t-test $p$$<$${0.01}$.}
    \vspace{0pt}
    \label{tab:results_ende}
\end{table}

\paragraph{Results on WMT'14 En${\rightarrow}$De.}
Table~\ref{tab:results_ende} lists several existing competitive NMT systems and ours. First, we can see that ``Multi-300k'' surpasses ``Transformer'' by +0.58 BLEU and +0.0101 Comet scores. Moreover, the BLEU and Comet scores of ``CBBGCA'' are respectively +1.02 and +0.0226 higher than ``Transformer'', verifying that CBKD at the second training stage brings further improvement.

Next, our proposed framework outperforms most recent competitive models. Specifically, ``CBBGCA'' yields a better result than ``ABD-NMT'' \cite{DBLP:conf/aaai/ZhangSQLJW18} and ``FKD-NMT'' \cite{DBLP:journals/taslp/ZhangXSL19} that only provide the NMT model with unidirectional future contexts. This proves the power of incorporating target-side bidirectional global context into the NMT model. Note that ``ABD-NMT'' needs two-pass decoding, which first obtains reverse state sequence by a backward decoder and then uses the forward decoder to generate final translations.
The only exception is ``SB-NMT'' \cite{DBLP:journals/tacl/ZhouZZ19} that designs elaborately cumstomized bidirectional decoding algorithms, which is actually not fairly comparable to ours because of its decoding manners and the involvement of synthetic training data.\footnote{The result of ``SB-NMT'' is the reported performance with checkpoint averaging technique from \cite{DBLP:journals/tacl/ZhouZZ19}.} These results demonstrate the effectiveness of our proposed training framework.

\begin{table}[!tbp]
    \centering
    \renewcommand{\arraystretch}{1.1}
    {
    \scalebox{0.90}{
    \begin{tabular}{ l | c | c | c | c }
    \toprule
    \hline
    \multirow{2}{*}{System} & \multicolumn{2}{c|}{Zh${\rightarrow}$En} & \multicolumn{2}{c}{En${\rightarrow}$Fr} \\ 
    \cline{2-5} & \textbf{BLEU} & \textbf{Comet} & \textbf{BLEU} & \textbf{Comet} \\ 
    \hline
    Transformer &  25.54 & 0.3187 &  40.97 & 0.5241 \\
    Multi-300k & 25.89$\ddagger$ & 0.3263$\ddagger$ & 41.26$\ddagger$ & 0.5271$\ddagger$ \\
    CBBGCA & \textbf{26.84}$\ddagger$ & \textbf{0.3544}$\ddagger$ & \textbf{41.54}$\ddagger$ & \textbf{0.5343}$\ddagger$\\
    \hline
    \end{tabular} }}
    \caption{BLEU (\%) and Comet scores on WMT'19 Zh${\rightarrow}$En and  WMT'14 En${\rightarrow}$Fr. `$\ddagger$': significantly ($p$$<$${0.01}$) better than the Transformer.} 
    \label{tab:results_chen&enfr}
\end{table}
\paragraph{Results on WMT'19 Zh${\rightarrow}$En and WMT'14 En${\rightarrow}$Fr.} From Table~\ref{tab:results_chen&enfr}, the ``Multi-300k'' preliminarily gains +0.35 and +0.29 BLEU scores over ``Transformer'' on Zh${\rightarrow}$En and En${\rightarrow}$Fr, respectively. Moreover, our ``CBBGCA'' further strongly outperforms ``Multi-300k'' and achieves a total of +1.30 and +0.57 BLEU score improvements over “Transformer” on the two datasets, respectively. In term of Comet, comparing different models, we can see similar results. Note that the sizes of WMT'19 Zh${\rightarrow}$En (20M) and WMT'14 En${\rightarrow}$Fr (36M) datasets are much larger than that of WMT'14 En${\rightarrow}$De (4.5M) dataset, demonstrating the effectiveness of our proposed framework on various language pairs.

\begin{table}[!tbp]
    \centering
    \renewcommand{\arraystretch}{1.1}
    {
    \scalebox{1.0}{
    \begin{tabular}{ l | c | c }
    \toprule
    \hline
    \textbf{System} &\textbf{BLEU}& \textbf{Comet} \\ 
    \hline
    Transformer & 27.30 & 0.2602 \\
    \hline
    Multi-300k & 27.88 & 0.2703 \\
    ~~~~ \textit{w/}. Dynamic & 27.97 & 0.2734 \\
    CBBGCA & 28.32 & 0.2828 \\
    ~~~~ \textit{w/o}. ShareEnc &  27.76 & 0.2683 \\
    ~~~~ \textit{w/o}. CBKD &  27.92 & 0.2712 \\
    \hline
    \end{tabular} }}
    \caption{BLEU (\%) and Comet scores of ablation results on the test set of WMT'14 En${\rightarrow}$De.} 
    \label{tab:results_ablation}
    \vspace{-10pt}
\end{table}

\section{Analysis}
\subsection{Ablation Study}
To fully investigate each part of our proposed training framework, we conduct ablation studies on WMT'14 En${\rightarrow}$De translation task. Table~\ref{tab:results_ablation} reports the ablation results on the test set. 

We first validate the necessity of our two-stage strategy by only training the model using the multi-task joint training. ``\textit{w/}. Dynamic'' means the weights in Equation~\ref{eq:stage1_objective} are dynamically adjusted. Specifically, we linearly increase $\lambda$ from $0.5$ to $1.0$ throughout the whole training process. We can see that its performance is just slightly higher than the fixed-weight ``Multi-300k'' and still significantly inferior to the two-stage ``CBBGCA''.

For the feasible options in two-stage strategy, ``\textit{w/o}. ShareEnc'' represents not sharing encoders at the first training stage and its performance decreases by $0.56$ BLEU score. This shows that the NMT encoder is enhanced by the joint training with the CMLM. As for ``\textit{w/o}. CBKD''\footnote{It differs from ``Multi-300k'' in that the CMLM is not optimized at the second stage (200k$\sim$300k steps) in ``\textit{w/o}. CBKD''. In ``Multi-300k'', between 200k$\sim$300k steps, the CMLM and NMT model continue to be jointly optimized by sharing their encoders.}, which means not performing KD on any target words at the second training stage (i.e., $\alpha$$=$$0$), its performance also decreases by $0.40$ BLEU score. This demonstrates the effect of pertinently incorporating bidirectional global context into the NMT model on its unconfidently-predicted target words.

\subsection{Effects of Different KD Strategies}
\vspace{5pt}
In our training framework, for each sentence pair, we adopt KD to transfer the knowledge of the CMLM into the NMT model only on the word set $\mathbf{y}_{m}$. The set is determined by masking $k$ target words whose NMT-predicted probabilities to the corresponding ground-truths are lower than a threshold $\epsilon$. Obviously, there are alternative strategies for the above process. Therefore, we further investigate the following variants:
\vspace{0pt}
\begin{itemize}
\setlength{\itemsep}{1pt}
\setlength{\parsep}{5pt}
\setlength{\parskip}{0pt}
\item \textbf{Random}: Regardless of confidence, we randomly select $k$ words of a target sentence to be masked for the CMLM.
\item \textbf{NMT-High}: 
As a contrast, we mask the target words whose NMT-predicted probabilities to the ground-truths are higher than $1-\epsilon$.
\item \textbf{NMT-Wrong}: 
We mask the target words where the predictions of the NMT model do not coincide with the corresponding ground-truths.
\item \textbf{All-at-Once}: In this variant, to validate the necessity of selectively distilling knowledge on a portion rather than all of target words, we generate CMLM-predicted probability distributions for all target words. As an extreme case, we mask all target words at once with only source sentences as input to the CMLM.
\item \textbf{Part-to-All}:
Instead of masking all target words at once, we generate the CMLM-predicted probability distributions in a part-to-all way. Concretely, we first generate the NMT-predicted probability distributions for all words. Then, all target words are divided into several non-overlapping subsets, each corresponding to a certain probability interval. For each time, we mask a subset of target words whose probabilities to the ground-truths are located within the corresponding interval. Particularly, because the hyper-parameter $\epsilon$ is $0.2$ in ``CBBGCA'', we have a total of $5$ iterations and the intervals are set to $[0.0,0.2]$, $[0.2,0.4]$, $[0.4,0.6]$, $[0.6,0.8]$, $[0.8,1.0]$.
\end{itemize}
Table~\ref{tab:results_strategy} lists the results with different KD strategies. We can observe that all these variants are inferior to our ``CBBGCA'' method. Particularly, the results of ``Random'' and ``NMT-High'' indicate that conducting knowledge distillation on either randomly selected or confidently-predicted target words is less effective than on those unconfidently-predicted ones. Next, the result of ``NMT-Wrong'' is lower than ``CBBGCA''. It may be due to the fact that the NMT model assigns low probabilities to some correctly-predicted target words.
Thus, the NMT model fails to absorb the beneficial knowledge from CMLM on these words. Lastly, ``All-at-Once'' and ``Part-to-All'' represent two approaches to generate CMLM-predicted probability distributions for all target words. It is reasonable for ``All-at-Once'' to obtain a a worse performance since the CMLM cannot predict well without any observable word on the target side. For ``Part-to-All'', we can see it improves the NMT model over ``Multi-300k'' but is still worse and takes more computational cost than ``CBBGCA''. This also echoes the finding in ``NMT-High'' that applying KD on confidently-predicted target words is not optimal. 
All these results demonstrate that it is crucial for the NMT model to pertinently exploit bidirectional global contexts on its unconfidently-predicted target words.

\begin{table}[!tbp]
    \centering
    \renewcommand{\arraystretch}{1.1}
    {
    \scalebox{1.0}{
    \begin{tabular}{ l | c | c }
    \toprule
    \hline
    \textbf{System} &\textbf{BLEU}& \textbf{Comet} \\ 
    \hline
    Transformer &  27.30 & 0.2602 \\
    Multi-300k &  27.88 & 0.2703 \\
    CBBGCA & \textbf{28.32} & \textbf{0.2828} \\
    \hline
    \multicolumn{3}{c}{KD Strategy} \\
    \hline
    Random &  28.09 & 0.2774 \\
    NMT-High &  27.85 & 0.2711 \\
    NMT-Wrong &  27.98 & 0.2734 \\
    All-at-Once &  27.64 & 0.2687 \\
    Part-to-All &  28.06 & 0.2769 \\
    \hline
    \end{tabular} }}
    \caption{BLEU (\%) and Comet scores on the test set of WMT'14 En${\rightarrow}$De with different KD strategies.} 
    \label{tab:results_strategy}
\end{table}
\vspace{-5pt}
\subsection{Change of Model Confidence}
We also investigate the change of model confidence with respect to target ground-truth words on the training set. Table~\ref{tab:conf_change} lists the percentage of tokens within each interval, in terms of NMT-predicted probability. Because the probability higher than $0.5$ must be the maximum across the vocabulary, we group $0.5$$\sim$$1.0$ as a whole high-confidence interval while the others are low-confidence intervals. 
From the table, we can observe that the number of tokens in low-confidence intervals drops. For instance, the number of tokens locating in $[0.0,0.2]$ becomes $0.69\%$ fewer, which is a notable change considering that the WMT'14 En${\rightarrow}$De training set contains roughly 4.5 million sentences with a total of approximately 140 million tokens.
This indicates that the NMT model becomes more confident about the target ground-truth words.
\vspace{0pt}
\begin{table}[!tbp]
    \centering
    \renewcommand{\arraystretch}{1.1}
    {
    \scalebox{0.62}[0.62]{
    \begin{tabular}{ c | c | c | c | c | c | c }
    \hline
    &[$0$,$0.1$)&[$0.1$,$0.2$)&[$0.2$,$0.3$)&[$0.3$,$0.4$)&[$0.4$,$0.5$)&[$0.5$,$1$]\\ 
    \hline
    Transformer & 25.67 & 6.19 & 4.82 & 4.44 & 4.54 & 54.34 \\
    \hline
    CBBGCA & 25.36 & 5.81 & 4.49 & 4.18 & 4.40 & 55.76 \\
    \hline
    $\Delta$ & -0.31 & -0.38 & -0.33 & -0.26 & -0.14 & +1.42 \\
    \hline
    \end{tabular} }}
    \caption{The percentage of tokens within each probability interval on the WMT'14 En${\rightarrow}$De training set. }
    \label{tab:conf_change}
\end{table}

\subsection{Integration with Large-scale PLM}
\vspace{5pt}
There are also some researches \cite{DBLP:conf/naacl/EdunovBA19,DBLP:conf/nips/ConneauL19,DBLP:conf/aaai/YangW0Z00020,DBLP:conf/aaai/WengYHCL20,DBLP:conf/acl/ChenGCLL20} that focus on incorporating large-scale PLMs into the NMT model. Different from these approaches that require pre-training on massive external data, in this work, the integration of CMLM into the training procedure is to directly provide the NMT model with target-side bidirectional global context without external data. To show that our proposed training framework is compatible and orthogonal to existing approaches involving large-scale PLMs, we conduct experiments with an external Roberta model ~\cite{DBLP:journals/corr/abs-1907-11692} in Appendix~\ref{sec:appendix_C}.

\subsection{Comparison over Stronger Systems}
\vspace{5pt}
To further validate our proposed training framework, we conduct experiments over stronger systems. Particularly, we first compare our proposed training framework with \cite{baziotis2020language} that used a bidirectional LM as prior to regularize the NMT model, which is similar to our method to some extent. Following their setting, we train a target-side language model using a 6-layer Transformer decoder. Then, it is used as the teacher model to impose soft constraints on the output of the NMT model. The upper rows of Table~\ref{tab:results_stronger} give the comparison results. We can see that our ``CBBGCA'' still outperforms ``Transformer + LM prior''. 

In addition, back-translated data are often used to boost NMT models.  We also involve additional back-translated data during training. Specifically, we use an in-house English-to-Chinese \textit{Transformer-base} model to translate the English sentences in the WMT'19 Zh${\rightarrow}$En training set. Then, we add these back-translated data to the WMT'19 Zh${\rightarrow}$En training set. The lower rows of Table~\ref{tab:results_stronger} lists the performance of our models under this setting. Similar to previous results, it shows that both of ``Multi-300k'' and ``CBBGCA'' consistently improve the NMT model on the BT-augmented WMT'19 Zh${\rightarrow}$En, demonstrating the the effectiveness of our proposed training framework.

\begin{table}[!tbp]
    \centering
    \renewcommand{\arraystretch}{1.1}
    {
    \scalebox{1.0}{
    \begin{tabular}{ l | c | c }
    \toprule
    \hline
    \textbf{System} &\textbf{BLEU}& \textbf{$\Delta$} \\ 
    \hline
    Transformer & 25.54 & ref. \\
    Transformer + LM prior & 25.85 & +0.31 \\
    CBBGCA & \textbf{26.84} & \textbf{+1.30} \\
    \hline
    \hline
    Transformer \textit{w/.} BT data & 25.85 & ref. \\
    Multi-300k \textit{w/.} BT data & 26.30 & +0.45 \\
    CBBGCA \textit{w/.} BT data & \textbf{27.12} & \textbf{+1.27} \\
    \hline
    \end{tabular} }}
    \caption{BLEU (\%) scores on the test set of WMT'19 Zh${\rightarrow}$En. The upper rows list the results of models trained using only the original training set of WMT'19 Zh${\rightarrow}$En. The lower rows represents the performances of models trained with additional back-translated data.} 
    \label{tab:results_stronger}
\end{table}

\subsection{Case Study}
In Appendix~\ref{sec:appendix_D}, we give an illustrative example on the WMT'19 Zh${\rightarrow}$En test set to show the improvements of our model.
\vspace{5pt}
\section{Related Work}
\vspace{5pt}
\subsection{Exploiting Global Context}
This line of research aims at modelling target-side global context in the reverse direction with an auxiliary model. \citet{DBLP:conf/naacl/LiuUFS16} first adopt L2R and R2L NMT models to independently generate translations through beam search, then re-rank the candidate list via their agreement. \citet{DBLP:conf/aaai/ZhangSQLJW18} employ a backward decoder to capture reverse target-side context, which is then exploited by the forward decoder. \citet{DBLP:conf/iclr/SerdyukKSTPB18} propose twin networks, where the forward network is encouraged to generate hidden states similar to those of the backward network. \citet{DBLP:journals/taslp/ZhangXSL19} present a future-aware knowledge distillation framework enabling the unidirectional decoder to explore the future context for word predictions. \citet{DBLP:journals/tacl/ZhouZZ19} propose a synchronous bidirectional NMT model with revised beam search algorithm that involves interactive L2R and R2L decodings. \citet{DBLP:conf/aaai/ZhangW0L0X19} also combine the L2R and R2L NMT models by considering their agreement, helping the model to generate sentences with better prefixes and suffixes. Although these methods indeed gain some improvements, the modelling of reverse global context is independent of the local context of preceding words. Meanwhile, they usually rely on elaborately designed mechanisms for burdensome multi-pass decoding.  

\subsection{KD from Pre-trained Language Model}
Another line of research is to exploit global contextual information contained in large-scale pre-trained language models (PLM) via knowledge distillation (KD). For example, \citet{DBLP:conf/naacl/EdunovBA19} and \citet{DBLP:conf/nips/ConneauL19} feed the top-layer representations of ELMo or BERT to NMT encoders. \citet{DBLP:conf/aaai/YangW0Z00020} explore three techniques to apply BERT on NMT models, namely asymptotic distillation, dynamic switch for knowledge fusion, and rate-scheduled updating. \citet{DBLP:conf/aaai/WengYHCL20} propose a training framework consisting of a dynamic fusion mechanism and a continuous KD paradigm to leverage the knowledge of various PLMs. \citet{baziotis2020language} incorporate a language model prior for low-resource NMT. \citet{DBLP:conf/acl/ChenGCLL20} fine-tune the BERT on the parallel corpus to make it aware of source input, and then utilize it to improve the NMT model via KD over all target words. Compared to this, CBBGCA jointly optimizes the NMT model and auxiliary model, leading to better performance. Moreover, our method just selectively conducts KD on a portion of target words, giving higher distillation efficiency, which is similar to~\cite{wang-etal-2021-selective}. Even though these PLM-based approaches have gained remarkable improvements, they unavoidably have some inherent limitations: (1) the monolingual PLMs lacks crucial bilingual information for translation; (2) the independence between PLM pre-trainings and NMT model training. 
In contrast, our proposed model is able to overcome these limitations.
\vspace{-1pt}
\section{Conclusion}
\vspace{0pt}
In this paper, we propose a CBBGCA training framework for the NMT model to effectively exploit target-side bidirectional global context with an auxiliary CMLM. The training consists of two stages. At the first stage, we introduce multi-task learning to benefit the NMT model by sharing its encoder with an auxiliary CMLM. Then, at the second stage, through confidence based knowledge distillation, we use the CMLM as teacher to especially refine the NMT model on unconfidently-predicted target words. 
Experimental results show that our framework can significantly improve the NMT model. Compared with previous work, neither external nor synthetic data are needed and only the NMT model is involved during inference.

\section*{Acknowledgements}
The project was supported by National Natural Science Foundation of China (No. 62036004, No. 61672440), Natural Science Foundation of Fujian Province of China (No. 2020J06001), and Youth Innovation Fund of Xiamen (No. 3502Z20206059). We also thank the reviewers for their insightful comments.

\bibliography{custom}

\begin{thebibliography}{37}
\expandafter\ifx\csname natexlab\endcsname\relax\def\natexlab#1{#1}\fi

\bibitem[{Bahdanau et~al.(2015)Bahdanau, Cho, and
  Bengio}]{DBLP:journals/corr/BahdanauCB14}
Dzmitry Bahdanau, Kyunghyun Cho, and Yoshua Bengio. 2015.
\newblock Neural machine translation by jointly learning to align and
  translate.
\newblock In \emph{ICLR 2015}.

\bibitem[{Baziotis et~al.(2020)Baziotis, Haddow, and
  Birch}]{baziotis2020language}
Christos Baziotis, Barry Haddow, and Alexandra Birch. 2020.
\newblock Language model prior for low-resource neural machine translation.
\newblock In \emph{EMNLP 2020}.

\bibitem[{Chen et~al.(2020)Chen, Gan, Cheng, Liu, and
  Liu}]{DBLP:conf/acl/ChenGCLL20}
Yen{-}Chun Chen, Zhe Gan, Yu~Cheng, Jingzhou Liu, and Jingjing Liu. 2020.
\newblock Distilling knowledge learned in {BERT} for text generation.
\newblock In \emph{ACL 2020}.

\bibitem[{Clark et~al.(2019)Clark, Luong, Khandelwal, Manning, and
  Le}]{clark2019bam}
Kevin Clark, Minh{-}Thang Luong, Urvashi Khandelwal, Christopher~D Manning, and
  Quoc~V Le. 2019.
\newblock Bam! born-again multi-task networks for natural language
  understanding.
\newblock In \emph{ACL 2019}.

\bibitem[{Conneau and Lample(2019)}]{DBLP:conf/nips/ConneauL19}
Alexis Conneau and Guillaume Lample. 2019.
\newblock Cross-lingual language model pretraining.
\newblock In \emph{NIPS 2019}.

\bibitem[{Devlin et~al.(2019)Devlin, Chang, Lee, and
  Toutanova}]{DBLP:conf/naacl/DevlinCLT19}
Jacob Devlin, Ming{-}Wei Chang, Kenton Lee, and Kristina Toutanova. 2019.
\newblock {BERT:} pre-training of deep bidirectional transformers for language
  understanding.
\newblock In \emph{NAACL-HLT 2019}.

\bibitem[{Edunov et~al.(2019)Edunov, Baevski, and
  Auli}]{DBLP:conf/naacl/EdunovBA19}
Sergey Edunov, Alexei Baevski, and Michael Auli. 2019.
\newblock Pre-trained language model representations for language generation.
\newblock In \emph{NAACL-HLT 2019}.

\bibitem[{Ghazvininejad et~al.(2019)Ghazvininejad, Levy, Liu, and
  Zettlemoyer}]{DBLP:conf/emnlp/GhazvininejadLL19}
Marjan Ghazvininejad, Omer Levy, Yinhan Liu, and Luke Zettlemoyer. 2019.
\newblock Mask-predict: Parallel decoding of conditional masked language
  models.
\newblock In \emph{EMNLP 2019}.

\bibitem[{Goodfellow et~al.(2016)Goodfellow, Bengio, and
  Courville}]{GoodBengCour16}
Ian~J. Goodfellow, Yoshua Bengio, and Aaron Courville. 2016.
\newblock \emph{Deep Learning}.
\newblock MIT Press, Cambridge, MA, USA.

\bibitem[{Hoang et~al.(2017)Hoang, Haffari, and Cohn}]{hoangHC17}
Cong Duy~Vu Hoang, Gholamreza Haffari, and Trevor Cohn. 2017.
\newblock Towards decoding as continuous optimisation in neural machine
  translation.
\newblock In \emph{EMNLP 2017}, pages 146--156.

\bibitem[{Kingma and Ba(2015)}]{DBLP:journals/corr/KingmaB14}
Diederik~P. Kingma and Jimmy Ba. 2015.
\newblock Adam: {A} method for stochastic optimization.
\newblock In \emph{ICLR 2015}.

\bibitem[{Koehn(2004)}]{DBLP:conf/emnlp/Koehn04}
Philipp Koehn. 2004.
\newblock Statistical significance tests for machine translation evaluation.
\newblock In \emph{EMNLP 2004}.

\bibitem[{Liu et~al.(2016)Liu, Utiyama, Finch, and
  Sumita}]{DBLP:conf/naacl/LiuUFS16}
Lemao Liu, Masao Utiyama, Andrew~M. Finch, and Eiichiro Sumita. 2016.
\newblock Agreement on target-bidirectional neural machine translation.
\newblock In \emph{NAACL-HLT 2016}.

\bibitem[{Liu et~al.(2019)Liu, Ott, Goyal, Du, Joshi, Chen, Levy, Lewis,
  Zettlemoyer, and Stoyanov}]{DBLP:journals/corr/abs-1907-11692}
Yinhan Liu, Myle Ott, Naman Goyal, Jingfei Du, Mandar Joshi, Danqi Chen, Omer
  Levy, Mike Lewis, Luke Zettlemoyer, and Veselin Stoyanov. 2019.
\newblock Roberta: {A} robustly optimized {BERT} pretraining approach.
\newblock \emph{CoRR}, abs/1907.11692.

\bibitem[{Meng and Zhang(2019)}]{meng2019dtmt}
Fandong Meng and Jinchao Zhang. 2019.
\newblock {DTMT:} {A} novel deep transition architecture for neural machine
  translation.
\newblock In \emph{AAAI 2019}.

\bibitem[{Miao et~al.(2021)Miao, Meng, Liu, Zhou, and
  Zhou}]{miao-etal-2021-prevent}
Mengqi Miao, Fandong Meng, Yijin Liu, Xiao-Hua Zhou, and Jie Zhou. 2021.
\newblock Prevent the language model from being overconfident in neural machine
  translation.
\newblock In \emph{ACL 2021}.

\bibitem[{Papineni et~al.(2002)Papineni, Roukos, Ward, and
  Zhu}]{DBLP:conf/acl/PapineniRWZ02}
Kishore Papineni, Salim Roukos, Todd Ward, and Wei{-}Jing Zhu. 2002.
\newblock Bleu: a method for automatic evaluation of machine translation.
\newblock In \emph{ACL 2002}.

\bibitem[{Peters et~al.(2018)Peters, Neumann, Iyyer, Gardner, Clark, Lee, and
  Zettlemoyer}]{DBLP:conf/naacl/PetersNIGCLZ18}
Matthew~E. Peters, Mark Neumann, Mohit Iyyer, Matt Gardner, Christopher Clark,
  Kenton Lee, and Luke Zettlemoyer. 2018.
\newblock Deep contextualized word representations.
\newblock In \emph{NAACL-HLT 2018}.

\bibitem[{Rei et~al.(2020)Rei, Stewart, Farinha, and
  Lavie}]{DBLP:conf/emnlp/ReiSFL20}
Ricardo Rei, Craig Stewart, Ana~C. Farinha, and Alon Lavie. 2020.
\newblock {COMET:} {A} neural framework for {MT} evaluation.
\newblock In \emph{EMNLP 2020}.

\bibitem[{Sennrich et~al.(2016)Sennrich, Haddow, and
  Birch}]{DBLP:conf/acl/SennrichHB16a}
Rico Sennrich, Barry Haddow, and Alexandra Birch. 2016.
\newblock Neural machine translation of rare words with subword units.
\newblock In \emph{ACL 2016}.

\bibitem[{Serdyuk et~al.(2018)Serdyuk, Ke, Sordoni, Trischler, Pal, and
  Bengio}]{DBLP:conf/iclr/SerdyukKSTPB18}
Dmitriy Serdyuk, Nan~Rosemary Ke, Alessandro Sordoni, Adam Trischler, Chris
  Pal, and Yoshua Bengio. 2018.
\newblock Twin networks: Matching the future for sequence generation.
\newblock In \emph{ICLR 2018}.

\bibitem[{Sohn et~al.(2015)Sohn, Lee, and Yan}]{NIPS2015_8d55a249}
Kihyuk Sohn, Honglak Lee, and Xinchen Yan. 2015.
\newblock Learning structured output representation using deep conditional
  generative models.
\newblock In \emph{NIPS 2015}.

\bibitem[{Song et~al.(2019)Song, Gildea, Zhang, Wang, and
  Su}]{DBLP:journals/tacl/SongGZWS19}
Linfeng Song, Daniel Gildea, Yue Zhang, Zhiguo Wang, and Jinsong Su. 2019.
\newblock Semantic neural machine translation using {AMR}.
\newblock \emph{Trans. Assoc. Comput. Linguistics}, 7.

\bibitem[{Su et~al.(2018)Su, Wu, Xiong, Lu, Han, and
  Zhang}]{DBLP:conf/aaai/SuWXLHZ18}
Jinsong Su, Shan Wu, Deyi Xiong, Yaojie Lu, Xianpei Han, and Biao Zhang. 2018.
\newblock Variational recurrent neural machine translation.
\newblock In \emph{AAAI 2018}.

\bibitem[{Su et~al.(2019)Su, Zhang, Lin, Qin, Yao, and
  Liu}]{DBLP:journals/ai/SuZLQYL19}
Jinsong Su, Xiangwen Zhang, Qian Lin, Yue Qin, Junfeng Yao, and Yang Liu. 2019.
\newblock Exploiting reverse target-side contexts for neural machine
  translation via asynchronous bidirectional decoding.
\newblock \emph{Artificial Intelligence}, 277.

\bibitem[{Sutskever et~al.(2014)Sutskever, Vinyals, and
  Le}]{DBLP:conf/nips/SutskeverVL14}
Ilya Sutskever, Oriol Vinyals, and Quoc~V. Le. 2014.
\newblock Sequence to sequence learning with neural networks.
\newblock In \emph{NIPS 2014}.

\bibitem[{Vaswani et~al.(2017)Vaswani, Shazeer, Parmar, Uszkoreit, Jones,
  Gomez, Kaiser, and Polosukhin}]{DBLP:conf/nips/VaswaniSPUJGKP17}
Ashish Vaswani, Noam Shazeer, Niki Parmar, Jakob Uszkoreit, Llion Jones,
  Aidan~N. Gomez, Lukasz Kaiser, and Illia Polosukhin. 2017.
\newblock Attention is all you need.
\newblock In \emph{NIPS 2017}.

\bibitem[{Wang et~al.(2021)Wang, Yan, Meng, and
  Zhou}]{wang-etal-2021-selective}
Fusheng Wang, Jianhao Yan, Fandong Meng, and Jie Zhou. 2021.
\newblock Selective knowledge distillation for neural machine translation.
\newblock In \emph{ACL 2021}.

\bibitem[{Watanabe and Sumita(2002)}]{DBLP:conf/coling/WatanabeS02}
Taro Watanabe and Eiichiro Sumita. 2002.
\newblock Bidirectional decoding for statistical machine translation.
\newblock In \emph{COLING 2002}.

\bibitem[{Weng et~al.(2020)Weng, Yu, Huang, Cheng, and
  Luo}]{DBLP:conf/aaai/WengYHCL20}
Rongxiang Weng, Heng Yu, Shujian Huang, Shanbo Cheng, and Weihua Luo. 2020.
\newblock Acquiring knowledge from pre-trained model to neural machine
  translation.
\newblock In \emph{AAAI 2020}.

\bibitem[{Yang et~al.(2020)Yang, Wang, Zhou, Zhao, Zhang, Yu, and
  Li}]{DBLP:conf/aaai/YangW0Z00020}
Jiacheng Yang, Mingxuan Wang, Hao Zhou, Chengqi Zhao, Weinan Zhang, Yong Yu,
  and Lei Li. 2020.
\newblock Towards making the most of {BERT} in neural machine translation.
\newblock In \emph{AAAI 2020}.

\bibitem[{Zhang et~al.(2016)Zhang, Xiong, Su, Duan, and
  Zhang}]{DBLP:conf/emnlp/ZhangXSDZ16}
Biao Zhang, Deyi Xiong, Jinsong Su, Hong Duan, and Min Zhang. 2016.
\newblock Variational neural machine translation.
\newblock In \emph{EMNLP 2016}.

\bibitem[{Zhang et~al.(2019{\natexlab{a}})Zhang, Xiong, Su, and
  Luo}]{DBLP:journals/taslp/ZhangXSL19}
Biao Zhang, Deyi Xiong, Jinsong Su, and Jiebo Luo. 2019{\natexlab{a}}.
\newblock Future-aware knowledge distillation for neural machine translation.
\newblock \emph{TASLP}, 27(12).

\bibitem[{Zhang et~al.(2020)Zhang, Zhou, Zhao, and
  Zong}]{DBLP:journals/ai/ZhangZZZ20}
Jiajun Zhang, Long Zhou, Yang Zhao, and Chengqing Zong. 2020.
\newblock Synchronous bidirectional inference for neural sequence generation.
\newblock \emph{Artificial Intelligence}, 281:103234.

\bibitem[{Zhang et~al.(2018)Zhang, Su, Qin, Liu, Ji, and
  Wang}]{DBLP:conf/aaai/ZhangSQLJW18}
Xiangwen Zhang, Jinsong Su, Yue Qin, Yang Liu, Rongrong Ji, and Hongji Wang.
  2018.
\newblock Asynchronous bidirectional decoding for neural machine translation.
\newblock In \emph{AAAI 2018}.

\bibitem[{Zhang et~al.(2019{\natexlab{b}})Zhang, Wu, Liu, Li, Zhou, and
  Xu}]{DBLP:conf/aaai/ZhangW0L0X19}
Zhirui Zhang, Shuangzhi Wu, Shujie Liu, Mu~Li, Ming Zhou, and Tong Xu.
  2019{\natexlab{b}}.
\newblock Regularizing neural machine translation by target-bidirectional
  agreement.
\newblock In \emph{AAAI 2019}.

\bibitem[{Zhou et~al.(2019)Zhou, Zhang, and Zong}]{DBLP:journals/tacl/ZhouZZ19}
Long Zhou, Jiajun Zhang, and Chengqing Zong. 2019.
\newblock Synchronous bidirectional neural machine translation.
\newblock \emph{Trans. Assoc. Comput. Linguistics}, 7:91--105.

\end{thebibliography}
\bibliographystyle{acl_natbib}

\appendix
\label{sec:appendix}
\newpage
\clearpage
\section*{Appendix}
\section{Datasets}
\label{sec:appendix_A}
\paragraph{WMT'14 En${\rightarrow}$De.} 
For English-to-German translation, the training set from the WMT 2014 contains about 4.5 million parallel sentence pairs. We use the \textit{newstest2013} and \textit{newstest2014} as the validation and test sets, respectively. We apply BPE to preprocess the data, obtaining a shared vocabulary of approximately 32,000 tokens. For evaluation, we adopt the case-sensitive BLEU scores by using the multi-bleu.perl script\footnote{https://github.com/moses-smt/mosesdecoder/blob/master/scripts/generic/multi-bleu.perl} and the Comet scores as our metrics.
\paragraph{WMT'19 Zh${\rightarrow}$En.}
For Chinese-to-English translation, we use the training data from the WMT 2019, which consists of about 20 million sentence pairs. The \textit{newstest2018} and \textit{newstest2019} are used as the validation and test sets, respectively. We also apply BPE to preprocess the data with 30,000 merge operations for both source and target languages. Then, we construct their corresponding vocabularies having roughly 47,000 Chinese tokens and 32,000 English tokens. To evaluate the models, we use cased BLEU scores calculated with the Moses mteval-v13a.pl script\footnote{http://www.statmt.org/moses/} as well as the Comet scores.
\paragraph{WMT'14 En${\rightarrow}$Fr.}
For English-to-French translation, the training set contains about 36 million parallel sentence pairs. We use the \textit{newstest2013} and \textit{newstest2014} as the validation and test sets, respectively. The preprocessing and evaluation are the same as those for WMT'14 En${\rightarrow}$De.
\section{Implementation Details}
\label{sec:appendix_B}
During training, we set dropout to 0.1 and use label-smoothing technique of value 0.1. For parameter updating, we employ the Adam optimizer \cite{DBLP:journals/corr/KingmaB14} with $\beta{1}=0.9$, $\beta{2}=0.998$ and $\epsilon$=$10^{-9}$. As for learning rate scheduling, we adopt the same strategy as \cite{DBLP:conf/nips/VaswaniSPUJGKP17} and set warm-up steps to 4,000. For all the three translation tasks, we train our models on Tesla V-100 GPU where we batch sentence pairs of similar lengths together containing roughly 32000 tokens. At the first stage, we train all models by sharing their encoders for 200,000 steps. At the second stage, we separate their encoders and fix the CMLM parameters. Then, the NMT model is solely optimized with the fixed CMLM by additional 100,000 steps, resulting in a total of 300,000 training steps.

At the inference time, we use beam search with the beam size 5. The results are reported with the statistical significance test \cite{DBLP:conf/emnlp/Koehn04} for both translation tasks.

\begin{table}[!tbp]
    \centering
    \renewcommand{\arraystretch}{1.1}
    {
    \scalebox{1.0}{
    \begin{tabular}{ l | c | c }
    \toprule
    \hline
    \textbf{System} &\textbf{BLEU}& $\Delta$ \\ 
    \hline
    \multicolumn{3}{c}{w/o Roberta} \\
    \hline
    Transformer &  27.30 & ref. \\
    Multi-300k &  27.88 & +0.58 \\
    CBBGCA & \textbf{28.32} & \textbf{+1.02} \\
    \hline
    \multicolumn{3}{c}{w/ Roberta} \\
    \hline
    Transformer &  27.64 & +0.34 \\
    Multi-300k &  28.13 & +0.83 \\
    CBBGCA & \textbf{28.68} & \textbf{+1.38} \\
    \hline
    \end{tabular} }}
    \caption{BLEU scores (\%) on the test set of WMT'14 En${\rightarrow}$De. The upper rows (w/o Roberta) represent the performance without using Roberta while the lower rows (w/ Roberta) are the results with the incorporation of the Roberta model using dynamic switch mechanism.} 
    \label{tab:results_dynamicswitch}
    \vspace{-5pt}
\end{table}

\begin{figure*}[!htb]
\centering
\includegraphics[scale=0.50]{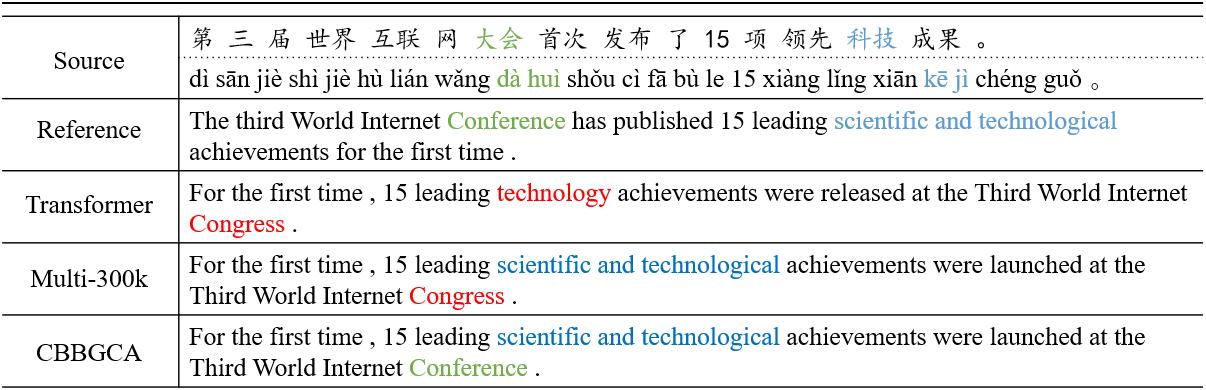}
\caption{An example on the WMT'19 Zh${\rightarrow}$En test set. The green and blue words are the source Chinese words and their corresponding English ground-truth translations. Those in red are wrongly translated words.
}
\label{fig:case}
\vspace{0pt}
\end{figure*}

\section{Integration with Large-scale PLM}
\label{sec:appendix_C}
In this experiment, we adopt the dynamic switching mechanism used in~\cite{DBLP:conf/aaai/YangW0Z00020} to incorporate externally pre-trained information into the NMT model. Specifically, we use Roberta \cite{DBLP:journals/corr/abs-1907-11692} as the large-scale PLM that provides our NMT model with external knowledge. Formally, a context gate is employed to control the amount of information flowing from the pre-trained model and our NMT model, which is computed as 
\begin{equation}
    g = \sigma(\mathbf{U}\mathrm{r}+\mathbf{V}\mathrm{h}),
\end{equation}
where $\sigma(\cdot)$ is the sigmoid function, ${\mathrm{r}}$ and $\mathrm{h}$ respectively represent the hidden states of each token obtained from the Roberta and the NMT encoder, $\mathbf{U}$ and $\mathbf{V}$ are the trainable parameter matrix. Then, the hidden states of the Roberta model and the NMT encoder are combined as
\begin{equation}
   \mathrm{h'} = g\odot\mathrm{r} + (1-g)\odot\mathrm{h},
\end{equation}
where $\odot$ is the operation of element-wise multiplication. 

Table~\ref{tab:results_dynamicswitch} lists the performance of our model with the integration of Roberta using dynamic switch mechanism. We can see that the incorporation of external Roberta brings some improvements to the NMT model under different settings (w/ Roberta vs. w/o Roberta). Besides, it is notable that ``CBBGCA'' still achieves consistent improvements over ``Transformer'' and ``Multi-300k'' regardless of whether using Roberta or not, which demonstrates the compatibility and orthogonality of our proposed training framework to those approaches~\cite{DBLP:conf/naacl/EdunovBA19,DBLP:conf/nips/ConneauL19,DBLP:conf/aaai/YangW0Z00020,DBLP:conf/aaai/WengYHCL20,DBLP:conf/acl/ChenGCLL20} of integrating external large-scale PLMs.

\section{Case Study}
\label{sec:appendix_D}
We conduct case study to illustrate the improvements of our model. Figure~\ref{fig:case} gives an example on the WMT'19 Zh${\rightarrow}$En test set with the outputs of different models. 

We can see that ``Transformer'' inappropriately translates the Chinese words ``\textit{dà huì}'' and ``\textit{kē jì}'' into ``\textit{Congress}'' and ``\textit{technology}'', respectively. For ``Multi-300k'', it correctly translates ``\textit{kē jì}'' into ``\textit{scientific and technological}''. Although it is somewhat acceptable for ``\textit{kē jì}'' to be translated as ``\textit{technology}'', ``\textit{scientific and technological}'' is a stricter translation and and more proper modifier considering that the succeeding  word``\textit{chéng guǒ}'' (``\textit{achievements}'' is a noun. The reason is that the multi-task joint training makes the NMT model aware of bidirectional global context rather than only the local context of preceding words. 

Moreover, ``CBBGCA'' successfully translates both ``\textit{dà huì}'' and ``\textit{kē jì}''. This indicates that the CBKD at the second training stage further gives our model more confidence to predict the ground truth ``\textit{Conference}'' rather than its near-synonym ``\textit{Congress}''. The above analyses show that our proposed training framework actually enhances the NMT model to capture bidirectional global context and significantly improves translation quality.


\end{document}